\documentclass[10pt,twocolumn,letterpaper]{article}

\usepackage{iccv}
\usepackage{times}
\usepackage{epsfig}
\usepackage{graphicx}
\usepackage{amsmath}
\usepackage{amssymb}

\usepackage[breaklinks=true,bookmarks=false]{hyperref}

\iccvfinalcopy 

\def\iccvPaperID{2532} 

\ificcvfinal\fi

\usepackage{subfigure} 
\usepackage{amssymb}
\usepackage{pifont}
\newcommand{\cmark}{\ding{51}}%
\newcommand{\xmark}{\ding{55}}%
\usepackage{bm}
\usepackage{booktabs}
\usepackage{caption} 
\usepackage[ruled,commentsnumbered]{algorithm2e}
\SetKwInOut{Parameter}{Paras.}

\begin{document}
	
	\title{Transductive Few-Shot Classification on the Oblique Manifold }
	
	\author{Guodong Qi $^{1,2}$, Huimin Yu $^{1,2,3}$, 	Zhaohui Lu $^{2}$, 	Shuzhao Li $^{1,2}$	\\
		$^{1}$College of Information Science and Electronic Engineering, Zhejiang University \\ $^{2}$ZJU-League Research \& Development Center, $^{3}$State Key Lab of CAD\&CG, Zhejiang University\\		
		{\tt\small \{guodong\_qi, yhm2005, leeshuz\}@zju.edu.cn,  lzh8910210@163.com }
	}
	
	\maketitle
	\ificcvfinal\thispagestyle{empty}\fi
	
	\begin{abstract}
		
		Few-shot learning  (FSL) attempts to learn with limited data. In this work, we perform the feature extraction in the Euclidean space and the geodesic distance metric on the Oblique Manifold (OM). Specially, for better feature extraction, we propose a non-parametric Region Self-attention with Spatial Pyramid Pooling  (RSSPP), which realizes a trade-off between the generalization and the discriminative ability of the single image feature. Then, we embed the feature to OM as a point. Furthermore, we design an Oblique Distance-based Classifier  (ODC) that achieves classification in the tangent spaces which better approximate OM locally by learnable tangency points. Finally, we introduce a new method for parameters initialization and a novel loss function in the transductive settings. Extensive experiments demonstrate the effectiveness of our algorithm and it outperforms state-of-the-art methods on the popular benchmarks: \textit{mini}-ImageNet, \textit{tiered}-ImageNet, 
		and Caltech-UCSD Birds-200-2011  (CUB).
		
	\end{abstract}
	
	\section{Introduction}
	
	Convolutional neural networks  (CNN) trained with large-scale labeled data have achieved the competitive performance as humans in recent years. However, different from these models struggling with a few labeled instances per class, humans learn rapidly by leveraging content and prior knowledge. To address this, few-shot learning  (FSL) has drawn increasing attention. FSL aims to learn a prior knowledge that can rapidly generalize to new tasks with limited samples.
	
	The \textit{transductive} learning methods~\cite{zikoLaplacianRegularizedFewShot2020,patacchiolaBayesianMetaLearningFewShot2020,liuEnsembleEpochwiseEmpirical2020,boudiafTransductiveInformationMaximization2020} and \textit{metric} learning methods~\cite{sung2018learning,kimFewshotVisualReasoning2020,yueInterventionalFewShotLearning2020} have been promising in a  recent line of works. Transductive learning has shown superior performance  over \textit{inductive} learning, because the unlabeled test examples are classified at once, instead of the one sample at a time as in inductive settings. Metric learning methods represent the image in an appropriate feature space and replace the fully connected layer in standard image classification~\cite{he2016deep} with the distance function, \eg Euclidean distance or \textit{cosine} distance. However, these works may lose the geometric inherent properties in the Euclidean space. Though it can be alleviated by data dimensionality reduction ~\cite{hotelling1933analysis,tenenbaum2000global,hinton2006reducing}, these methods are not generalized well to new tasks or easy to over-fit in the few-shot settings.

	Another solution is to make use of Riemannian geometry~\cite{eisenhart1997riemannian}. Riemannian geometry studies real smooth differential manifolds and it defines several geometric notions, \eg the length of a curve, with a Riemannian metric. Admitting the Riemannian geometry, the Grassmannian manifold~\cite{huangProjectionMetricLearning2015, wangGraphEmbeddingMultiKernel2021} and the SPD manifold~\cite{nguyenNeuralNetworkBased2019} are highly prevalent in modeling characters of image sets and videos, where \textit{intra-class} variance, \eg, illumination conditions or other scenarios, are comprised. They are capable of ``filling-in'' missing images. However, the advantages are based on modeling sufficient number of images for each class. It is infeasible to apply the aforementioned manifolds to FSL. 
	
	Another manifold that admits Riemannian geometry, is the \textit{oblique manifold}  (OM)~\cite{trendafilovMultimodeProcrustesProblem2002}, which is an embedded submanifold with all normalized columns and it is used for independent component analysis~\cite{absilJointDiagonalizationOblique2006}. We argue that OM is superior in FSL. The reasons are in two-fold: 1) \textit{Whitening} is not required in OM. The whitening step is to estimate the covariance matrix and remove the scaling indeterminacy of data in Grassmannian manifold or SPD manifold, which is infeasible to perform whitening step for a single image classification. Without whitening, OM is free of extrinsic constraints. 2) All the columns of OM have unit Euclidean norm. OM offers an intrinsic property similar to the \textit{L2-normalization} in Euclidean space. In this way, the CNN features in Euclidean space can be embedded to OM more easily. 
	
	However, the absence of whitening may lead to weak generalization of OM, since the whitening step is to remove the impact of intra-variance of data. To address this, we resort to the powerful CNN to enhance the generalization, as generalization emerged from the pretrained CNN~\cite{he2016deep,zhangFewShotLearningSaliencyGuided2019,manglaChartingRightManifold2020}, ensemble learning~\cite{dvornikDiversityCooperationEnsemble2019} or SPP~\cite{he2015spatial}. However, more generalization means less discriminative CNN~\cite{pautratOnlineInvarianceSelection2020}. Recent works ~\cite{chen2017sca, woo2018cbam} show that discriminative local regions can be enriched when training with self-attention networks. Inspired by these works, to improve the generalization of OM without losing the discriminative representation, we propose a non-parametric \textit{region self-attention with spatial pyramid pooling}  (RSSPP).  Specially, given a CNN feature map from a single image, RSSPP applies multi-kernel max-pooling similar to SPP~\cite{he2015spatial} to enhance the generalization of features. Then, RSSPP employs the self-attention mechanism to improve the discriminative ability. Note that our RSSPP is non-parametric and avoids over-fitting in FSL.

	\begin{figure}[]
		\centering
		
		\subfigure[Forward]{
			\includegraphics[width=.2\textwidth]{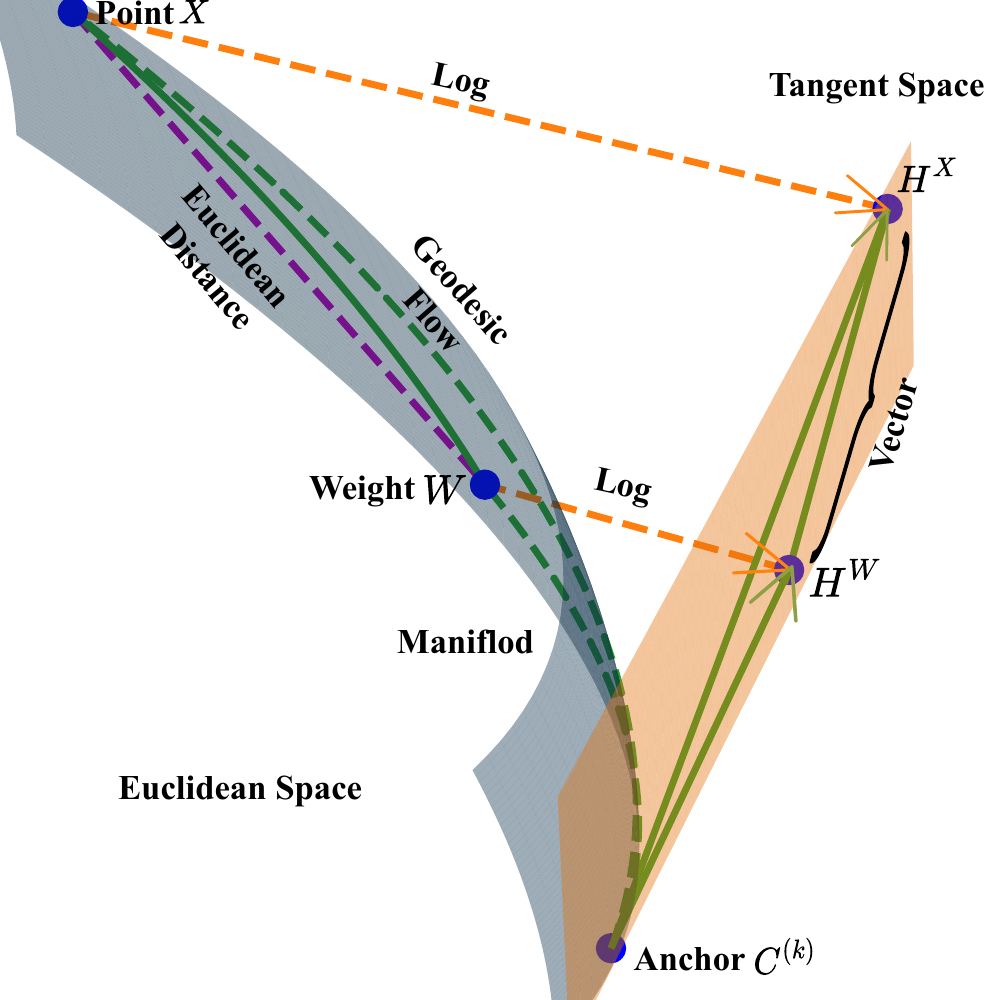}
			\label{fig_into_a}
		}
		\subfigure[Backward]{
			\includegraphics[width=.2\textwidth]{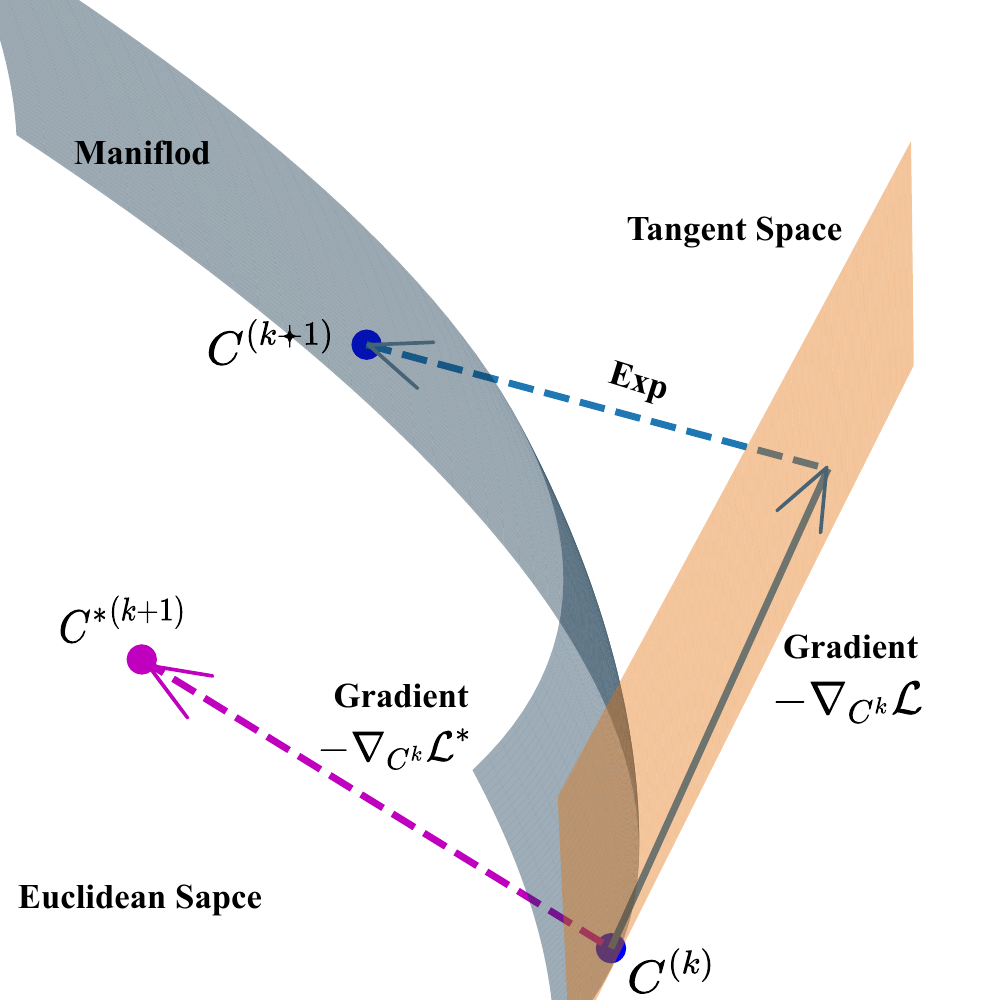}
			\label{fig_into_b}
		}	
		
		\caption{Illustration of Oblique Distance-based Classifier (ODC), which is parameterized with \textit{Weights} and \textit{Anchors}.   (a) Forward: more precious than Euclidean distance, the geodesic flow is transformed to a vector in the tangent space by $\operatorname{Log}$.  (b) Backward: parameters in Euclidean space are updated in the directions of negative gradient, while manifold-valued parameters are updated along the manifold by $\operatorname{Exp}$. Best viewed in color.}
		\label{fig_intro_manifold}
	\end{figure}

	After applying RSSPP, to take advantage of the aforementioned properties of OM, we map the Euclidean features to embedded features\footnote{We also refer to the embedded features as \textit{manifold-valued features} or \textit{points} to be distinguished from features in Euclidean.} on OM. 
	Since distance-based classifier is widely used in FSL \cite{boudiafTransductiveInformationMaximization2020,tianRethinkingFewShotImage2020} , we present a novel \textit{oblique distance-based classifier} (ODC) to classify the manifold-valued features. As illustrated in Figure \ref{fig_intro_manifold}, ODC is parameterized with \textit{weights} and \textit{anchors}. Both are members of OM, where the weights put emphasis on classifying points by the precise geodesic distance, and the anchors are points of tangency, which determine the tangent spaces. The tangent spaces offer Euclidean properties, so we can employ modern machine learning tools to perform classification with an iterative way. In the forward pass, the geodesic distance is transformed to the vector in tangent spaces at the anchors with the logarithmic map  ($\operatorname{Log}$). During the backward pass, the gradient is computed, mapped back to the geodesic flow by exponential map ($\operatorname{Exp}$), and used to update the parameters without leaving manifold.

	As the representation ability of tangent spaces decreases a lot when points are too far from anchors~\cite{souzaInterfaceGrassmannManifolds2020}, the anchors need to be initialized appropriately. We argue that the anchors should be in the neighborhood of the Karcher mean (KM)~\cite{grove1973conjugatec} among associated points. Considering the potential ill-distribution gap between the train samples and the test samples, we initialize the anchors  by selecting samples in the transductive settings, and design a weighted loss function to integrate the tangent spaces. Furthermore, considering the calculation of KM is an NP-hard problem, we propose a pseudo-KM method to utilize the weighted mean (\ie inner product) operation on the features in Euclidean space, and then embed the mean feature to OM. Empirically we observe the initialization of weights is also critically important. Similar to the anchors, we calculate prototypes~\cite{snellPrototypicalNetworksFewshot2017} and embed the prototypes to OM as the initial value of weights.  
	
	Finally, we acquire the classification scores by performing weighted sum over $\operatorname{softmax}$ on the Euclidean distance in tangent space. The effectiveness of our method is demonstrated by extensive experiments on multiple datasets. To conclude, our main contributions are summarized as follows: 
	\begin{itemize}
		\item To our best knowledge, it is the first time to model FSL on OM, which intrinsically satisfies the normality constraint and without the need for whitening. 
		
		\item We propose a non-parametric RSSPP. It applies the multi-kernel max-pooling similar to SPP to enhance generalization and the self-attention mechanism to improve discriminative ability.
		
		\item To perform classification in the oblique manifold, we propose the ODC parameterized with weights and anchors. ODC is initialized with a carefully designed strategy. We also design a weighted sum loss function over the anchors, and utilize the exponential map to update the weights and anchors.
		
		\item The experiments on popular datasets demonstrate that  our algorithm on FSL significantly outperforms the baseline and achieves new state-of-the-art performance on all of them.
		
	\end{itemize}

	\section{Related Work}

	{\bf Few-Shot Learning.}  
	Metric learning approaches~\cite{sung2018learning,lifchitz2019dense, kimFewshotVisualReasoning2020,yueInterventionalFewShotLearning2020,zhangDeepEMDFewShotImage2020}  and optimization-based approaches~\cite{finn2017model,li2019learning,park2019meta,rajeswaran2019meta,leeMetaLearningDifferentiableConvex2019} are two common line algorithms in few-shot learning. Optimization-based methods adapt the model parameters to new tasks rapidly with the inner loop.  Metric learning methods, which are more related to ours, target at learning a good embedding in appropriate spaces and distance function to metric the relationship of feature. These methods refer to the whole image~\cite{vinyals2016matching, snellPrototypicalNetworksFewshot2017} or local regions~\cite{zhangDeepEMDFewShotImage2020, dvornikSelectingRelevantFeatures2020} as embedding. However, they all compute distance in the Euclidean space. Instead, we utilize the geodesic distance in the oblique manifold.
	
	Recently, fine-tuning methods~\cite{dhillonBaselineFewShotImage2020,tianRethinkingFewShotImage2020} have shown that a pretrained in base datasets also provide a good embedding for news tasks, and they only fine-tune the last classifier layer. Transductive learning~\cite{liuLEARNINGPROPAGATELABELS2019,qiaoTransductiveEpisodicWiseAdaptive2019,yangDPGNDistributionPropagation2020,zikoLaplacianRegularizedFewShot2020} is another technique that classifies unlabeled data at once, instead of the one-by-one sample in inductive methods. TIM~\cite{boudiafTransductiveInformationMaximization2020} maximizes the mutual information between the unlabeled data and the label probability together with the supervised loss during fine-tuning. Based on Tim, we propose a novel weighted sum loss function.
	
	{\bf Manifold Learning. } 	Manifold learning, including dimensionality reduction~\cite{ hotelling1933analysis, tenenbaum2000global, hinton2006reducing} and Riemannian manifolds learning ~\cite{harandiGraphEmbeddingDiscriminant2011,gopalanDomainAdaptationObject2011,gongGeodesicFlowKernel2012,huangProjectionMetricLearning2015,montiGeometricDeepLearning2017,weiLearningDiscriminativeGeodesic2018} has attracted significant attention in the past few years.   Popular manifolds, \eg Grassmannian manifolds~\cite{chakrabortyManifoldNetDeepNeural2020,wangGraphEmbeddingMultiKernel2021} and SPD manifolds~\cite{huang2017riemannian,nguyenNeuralNetworkBased2019}, leverage modern Euclidean machine learning tools and maintain the structure of the manifolds meanwhile. Recently, Souza \etal~\cite{souzaInterfaceGrassmannManifolds2020} propose to transform the manifold points to tangent vectors at random tangency points by logarithmic map. However, they all focus on image sets and video data, while we solve the single image classification in the oblique manifold. Furthermore, different from Souza \etal~\cite{souzaInterfaceGrassmannManifolds2020}, the tangency points in our method are defined carefully.
	
	\begin{figure*}[!htbp]
		\centering
		\includegraphics[width=.9\textwidth]{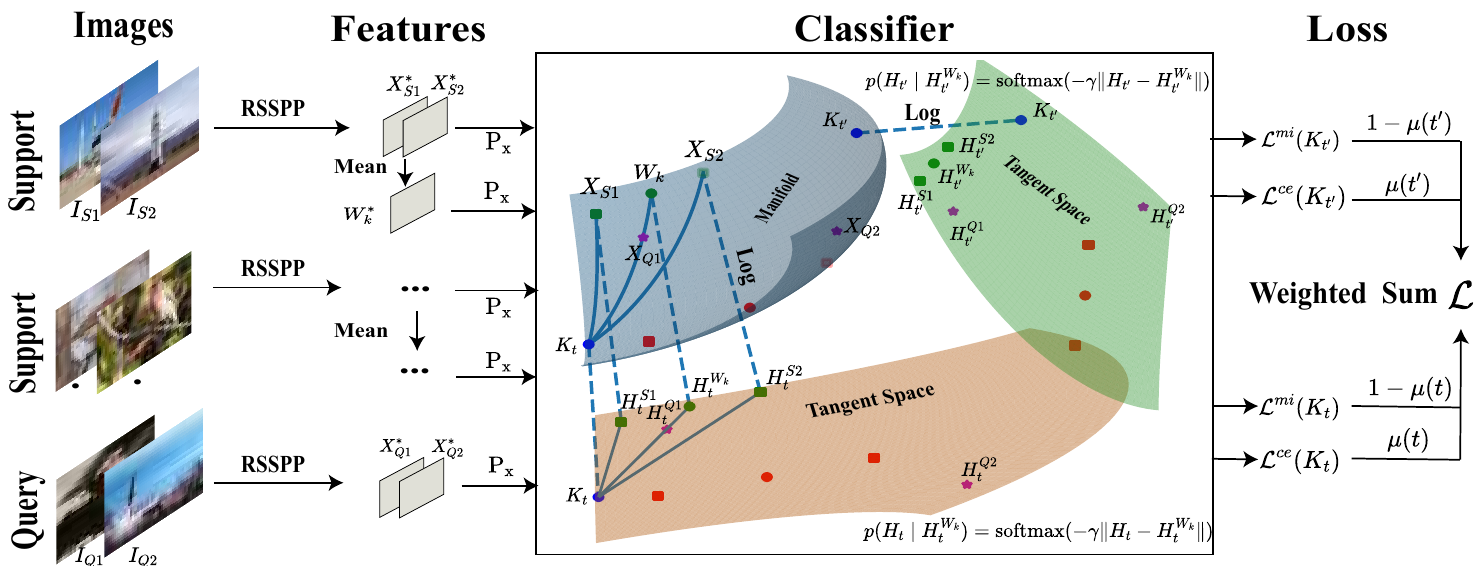}
		\caption{The overall pipeline of our method. For clarity and visibility, we take 2-way 2-shot classification as an example, plot partially, and push the tangent spaces away from the manifold in the classifier. Given images $\{ \bm{I}_{S}, \bm{I}_{Q}\}$, we obtain features $\{ \bm{X}_S^*, \bm{X}_Q^* \}$ by RSSPP and the prototypes $\bm{W}_k^*$. Then, they are embedded as points $\{ \bm{X}_S, \bm{X}_Q, \bm{W}_k\}$  by $\operatorname{P_X}$. Next, the geodesic flows between points and anchors $\{ \bm{K}_t$, $\bm{K}_{t^{'}} \}$ are transformed to tangent vectors $ \{ \bm{H}^S_t , \bm{H}^Q_t, \bm{H}_{t}^{W_k}, \cdots \} $ in the tangent spaces by $\operatorname{Log}$. Finally, we calculate the probability distribution by $\operatorname{softmax}$ and total loss $\mathcal{L}$ for fine-tuning.}
		\label{fig_overview}
	\end{figure*}

	{\bf Other Related Work. } 
	Descriptors, \eg, modern CNN~\cite{he2016deep}, have been utilized to encode an image to capture as much generalization as possible in recent years. Self-attention~\cite{chen2017sca,fu2019dual}, \eg, spatial attention~\cite{Wang_2018_CVPR} and channel attention~\cite{tan2019mnasnet}, are applied in computer vision as a complement to the whole features~\cite{woo2018cbam,hu2019local,zhaoExploringSelfAttentionImage2020} to enrich discriminative information. In this work, we employ SPP~\cite{he2015spatial} and spatial attention to make a trade-off between generalization and discriminative ability. Different from them, no extra parameters in our method ensures to avoid over-fitting.

	\section{Preliminaries}
	Before presenting the methods, we give a brief overview of the oblique manifold geometry and the few-shot problem.

	\subsection{Oblique Manifold}
	The \textit{oblique manifold}  (OM) $\mathcal{O} (n,p)$ is defined as a set of matrix in $\mathbb{R}^{n\times p}$ with unit Euclidean norm columns~\cite{trendafilovMultimodeProcrustesProblem2002,absilJointDiagonalizationOblique2006}. Formally, OM is defined by:
	\begin{equation}\label{eq1}
		\mathcal{O} (n,p)= \left\{ \bm{Y}  \in {\mathbb{{R}}}^{n\times p}: \operatorname{diag} (\bm{Y}^T\bm{Y} )=\bm{I_p} \right\} 
	\end{equation}
	where $\operatorname{diag} (\cdot)$ denotes the diagonal matrix.
	
	\subsection{Tangent Spaces}
	A tangent space $T_{\bm{K}} \mathcal{O}$ is the set of all tangent vectors to $\mathcal{O}$ at point $\bm{K}$. It provides a local vector space approximation of the manifold. Any tangent vector $\dot{K}$ must meet this constraint:  $ {\dot{\bm{K}}}^T\bm{K}+\bm{K}^T\dot{\bm{K}}=\bm{0} $. To be exact, the tangent space at $\bm{K}$  on OM is calculated by: 
	\begin{equation}\label{eq2}
		T_{\bm{K}} \mathcal{O} (n, p) = \{ \dot{\bm{K}} \in \mathbb{R}^{n\times p}
		: \operatorname{diag} (\bm{K}^T\dot{\bm{K}}) = \bm{0} \}
	\end{equation} 
	
	\subsection{Exponential and Logarithmic Maps}
	Given a length $t$ on a geodesic $\gamma (t)$, where $\gamma (0)$ is a start point, $\dot{\gamma} (t)$ is a direction, the specific points are obtained by the exponential map $\operatorname{ Exp}:\mathcal{O} \times T\mathcal{O} \times \mathbb{R} \rightarrow \mathcal{O} $. Specially, we denote $\gamma (t)=\operatorname{Exp}_{\bm{K}}\bm{H} $,  which shows point $\gamma (t)$ in geodesic emanating by:
	\begin{equation}\label{eq-exp_map}
		\operatorname{Exp}_{\bm{K}} (\bm{H}) = \bm{K} \cos (t\| \dot{\bm{K}} \|)+\frac{\dot{\bm{K}}}{\|\dot{\bm{K}}\|}\sin (t\| \dot{\bm{K}} \| )
	\end{equation}
	$ \| \cdot \|$ refers to the Frobenius norm in the ambient space, $\bm{H} = t\dot{\gamma} (0)$, $\gamma (0)=\bm{K}$ and $\dot{\gamma} (0)=\dot{\bm{K}}$
	
	Logarithmic map is the inverse of exponential map. Given two points $\bm{K}$ and $\bm{X}$ on OM, we denote $\bm{H}=  \operatorname{Log}_{\bm{K}}{\bm{X}}$ where $\operatorname{Log}:\mathcal{O}\times \mathcal{O} \rightarrow T\mathcal{O}$. It outputs the tangent vector $\bm{H}$ at $\bm{K}$ pointing towards $\bm{X}$, describing the shortest path curve from $\bm{K}$ to $\bm{X}$. Formally, $\operatorname{Log}$ is written as:
	\begin{align}\label{eq-log_map}
		\operatorname{dist} (\bm{K}, \bm{X}) &= \sqrt{\sum_{i=1}^p \arccos^2 (\operatorname{diag}  (\bm{K}^T \bm{X}))_i}  \notag \\
		\operatorname{P_{{k}}} (\bm{X}-\bm{K}) &=  (\bm{X}-\bm{K}) - \bm{K}\operatorname{diag} (\bm{K}^T  (\bm{X}-\bm{K})) \notag \\
		\operatorname{Log}_{\bm{K}} (\bm{X}) &= \frac{\operatorname{dist} (\bm{K}, \bm{X})}{\|P_{\bm{K}} (\bm{X}-\bm{K})\|}\operatorname{P_{{k}}} (\bm{X}-\bm{K})
	\end{align}
	
	\subsection{Few-Shot Learning}

	Given a \textit{base} training set $\mathbb{D}_{base}:=\{ \bm{I}_i,{y}_i: \bm{I}_i \in \mathbb{I}_{base}, {y}_i \in \mathbb{C}_{base} \}_{i=1}^{N_{base}}$, $ \bm{I}_i$ denotes raw image sample $i$ from base image set $\mathbb{I}_{base}$ and ${y}_i$ is its associated label in the base training set $\mathbb{C}_{base}$. In the few-shot scenario, given a \textit{novel} test set $\mathbb{D}_{novel}:=\{\bm{I}_i,{y}_i: \bm{I}_i \in \mathbb{I}_{novel}, {y}_i \in \mathbb{C}_{novel} \}_{i=1}^{N_{novel}}$,  where $\mathbb{D}_{base} \cap \mathbb{D}_{novel} = \phi $, we create the $c$-way $k_S$-shot  \textit{tasks} randomly sampled with a few number of examples. Specially, we  sample $c$ classes from $\mathbb{C}_{novel}$  and for each sampled class we choose $k_S$ samples randomly. These selected samples form \textit{support} set $S$ with size $|S|:=k_S \times c$. Similarity, we obtain  \textit{query} set $Q$ with size $|Q|:=k_Q \times c$  by sampling $k_Q$ unlabeled  (unseen) examples for each classes randomly.

	\section{Method}
	In this section, we firstly present the detail of RSSPP. Then, we describe the embedded method on OM. Next, we design ODC and show the  strategy for initializing parameters. Finally, we reveal the loss function and optimization on OM. The overview of our method is illustrated in Figure \ref{fig_overview}.

	\subsection{Region Self-Attention with Spatial Pyramid Pooling}
\begin{figure}[t]
		\centering
		\includegraphics[width=.45\textwidth]{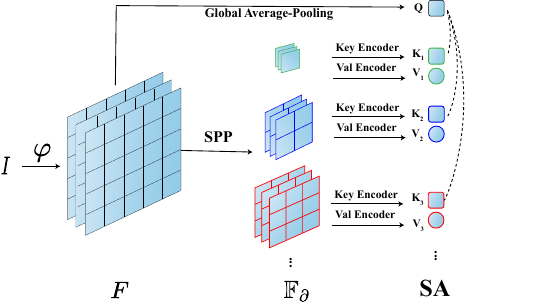}
		\caption{The overview of Region Self-attention with Spatial Pyramid Pooling. Both key and value encoders are non-parametric.}
		\label{fig_RSSPP}
	\end{figure}

	To make a trade-off \cite{pautratOnlineInvarianceSelection2020} for features between generalization and discrimination from a single image, we offer the non-parametric region self-attention with spatial pyramid pooling  (RSSPP). As illustrated in Figure \ref{fig_RSSPP}, given an image $\bm{I}$ and a pre-trained CNN $\varphi$ from the base datasets, we obtain the image features $\bm{F} \in \mathbb{R}^{n \times h \times w}$ by $\bm{F}:=\varphi (\bm{I})$, where $n$, $h$, and $w$ denote the feature dimension, the spatial height and width of the feature map respectively. To get the generalization feature set $\mathbb{F}_{\partial}$, the RSSPP utilizes multi-kernel max-pooling similar to SPP:
	\begin{align}\label{eq_mp}
		\bm{s}_i &:= (\lfloor h / i \rfloor, \lfloor w / i \rfloor) \notag \\
		\bm{k}_i &:=  (h- (i+1)\cdot\lfloor h / i \rfloor, w- (i+1)\cdot\lfloor w / i \rfloor)   \notag \\
		\mathbb{F}_{\partial} &:= \{ \bm{F}_i = \operatorname{MP} (\bm{F}, \bm{k}_i, \bm{s}_i)\in \mathbb{R}^{n\times i\times i} \}_{i=1}^p 
	\end{align}
	where $\operatorname{MP} (\cdot)$ denotes the max-pooling operator, $p$ is the number of kernels satisfying $p < \min (h,w)$, $\bm{k}_i$ is the max-pooling kernel size, $\bm{s}_i$ is the stride, and $\lfloor\cdot\rfloor$ is the largest integer that is less than or equal to $\cdot$. Next, to improve discriminative ability, we utilize the self-attention mechanism and denote the key encoder, value encoder and query as:
	\begin{align}\label{eq_encoder}
		\bm{k_i} &:= \frac{\operatorname{GAP} (\bm{F}_{i}) \cdot \operatorname{GAP} (\bm{F}_{p})}{\operatorname{GAP} (\bm{F}_{1})}  + \operatorname{GAP} (\bm{F}_{i}) \notag \\
		\bm{v_i} &:= \operatorname{GAP} (\bm{F}_{i}) \notag \\
		\bm{q} &:= \operatorname{GAP} (\bm{F}_{p})
	\end{align} where $\bm{k_i}, \bm{v_i}, \bm{q} \in \mathbb{R}^n$, $\bm{F}_{i} \in\mathbb{F}_{\partial}$ and $\operatorname{GAP}$ is the global average-pooling.
	To the end, the key encoder enhances the local discriminative region furthermore, the value encoder encodes the current feature map and the query encoder stands for applying the global average pooling on the whole feature map. Finally, following the self-attention and the residual shortcut  (SA), we obtain the features $\bm{x}^*_i$:
	\begin{equation}\label{eq_sa}
		\bm{x}^*_i :=\operatorname{SA} (\bm{q},\bm{k}_i, \bm{v}_i):=  \operatorname{softmax} (\frac{\bm{q}\bm{k}_i^T}{\sqrt{n}})\bm{v}_i + \bm{v}_i
	\end{equation}	
	Finally, we collect the feature matrix $\bm{X}^{*} \in \mathbb{R}^{n\times p}$ by $\bm{X}^{*} := \operatorname{Cat} (\{\bm{x}_i^* \}_{i=1}^p)$, where $\operatorname{Cat} (\cdot)$ denotes concatenating the given sequence of features. For simplicity, we formulate the series connected functions in this section as $\operatorname{RSSPP}$:
	\begin{equation}\label{eq-RSSPP}
		\operatorname{RSSPP}:=\operatorname{Cat} \circ \operatorname{SA} \circ \operatorname{MP} \circ \varphi 
	\end{equation}
	In FSL, we obtain the support features set $ {\mathbb{S}^{*}}:=\{\bm{X}^{*}_S =\operatorname{RSSPP} (\bm{I}) \in \mathbb{R}^{n\times p}\}_{\bm{I} \in S}$ and the query features set $\mathbb{Q}^{*}:=\{\bm{X}^{*}_Q =\operatorname{RSSPP} (\bm{I}) \in \mathbb{R}^{n\times p}\}_{\bm{I} \in Q}$.
	
	\subsection{Embedded on the Oblique Manifold}
	
	Since $\operatorname{diag} ({\bm{X}^{*}}^T\bm{X}^{*}) \neq \bm{I_p}$, the feature matrix $\bm{X}^{*}$ is not a member of OM.  We apply a projector $\operatorname{P_x} (\cdot)$ to get the manifold-valued feature matrix $\bm{X}$: 
	\begin{equation}\label{eq_projector}
		\bm{X}:=\operatorname{P_x} (\bm{X}^{*}) = \operatorname{Cat} ( \{\frac{\bm{x}^{*}_i}{\|\bm{x}^{*}_i\|}\}_{i=1}^p )
	\end{equation}
	In this way, the support embedding set and query embedding set are collected as $\mathbb{S}:= \{\bm{X}_S = \operatorname{P_x} (\bm{X}_{S}^{*})\in \mathbb{R}^{n\times p}\}$ and  $\mathbb{Q}:= \{\bm{X}_Q = \operatorname{P_x} (\bm{X}_{Q}^{*})\in \mathbb{R}^{n\times p}\}$ respectively. 
	
	\subsection{Classification on the Oblique Manifold}

	In this work, we define the oblique distance-based classifier  (ODC) with weights $  \mathbb{W} := \{ \bm{W}_{k} \in \mathbb{R}^{ n \times p } \}_{k=1}^c $ and anchors  $\mathbb{K}:= \{\bm{K}_t\}_{t=0}^{\tau} $, where $\tau +1$ is the length of the anchors set. Both weights and anchors are points on OM. 
	
	\textbf{Initialize Anchors. }	 ODC seeks the anchor $\bm{K}_{t} $ so that the tangent vector $ \bm{H}_t=\operatorname{Log}_{\bm{K}_{t}} (\bm{X})$ is as representative as possible. Commonly, $\operatorname{Log}$ is a helpful approximation of manifold in a local neighborhood of the anchor ${\bm{K}_t}$, however, its representation capability decreases to points too far from $\bm{K}_{t} $. Thus, it is critical to initialize proper anchors. 
	
	Intuitively, ${\bm{K}_t}$ are required to be close to the Karcher mean  (KM)~\cite{grove1973conjugatec} of associated points. However, the computation of exact KM is an NP hard problem. Instead, we propose the pseudo-KM as initial value of ${\bm{K}_t}$. In particular, considering $\bm{X}^{*}$ in Euclidean space, it is reasonable to calculate Euclidean anchors $\bm{K}^*_0$ by weighted mean over the support features:
	$ \bm{K}^*_0 := \frac{1}{\left|S\right|}\sum_{\bm{X}_S^* \in \mathbb{S}^{*}} \bm{X}_S^* $. Then, we get manifold anchors $\bm{K}_0$: $\bm{K}_0 = \operatorname{P_x} (\bm{K}^*_0)$.
	
	Furthermore, considering the potential ill-distribution gap between  the support samples and  the query samples in FSL, we initialize anchors with a strategy that the associated samples are selected carefully in the transductive settings. Formally, the general initial $\bm{K}_t$ is computed by: 
	\small	
	\begin{equation}\label{eq_anchor_k} \scriptscriptstyle 
		\bm{K}_{t} := 
		\begin{cases}
			\operatorname{P_x} (  \frac{\sum_{\bm{X}_S^* \in \mathbb{S}^{*}} \bm{X}_S^* } { |S|}) & \tau=0 \\
			\\
			\operatorname{P_x} (  \frac{ (\tau-t)  \cdot \sum_{\bm{X}_S^* \in \mathbb{S}^{*}} \bm{X}_S^* + t \cdot \sum_{\bm{X}_Q^* \in \mathbb{Q}^{*}} \bm{X}_Q^* } { (\tau-t)  \cdot |S| + t \cdot |Q|})	& \tau >0 \\
		\end{cases}
	\end{equation}	
	\normalsize

	\begin{algorithm}[]
		\small
		\Parameter{Number of Anchors $\tau$, Number of Kernels $p$, Iterations $iter$, Learning rate $\lambda_r$, Temperature $\gamma$, Weights factor $\{\lambda, \alpha, \mu(t)\}$ }
		\KwIn{Pre-trained CNN $\varphi$. Task $\{S, Q\}$}
		
		Compute $\{\mathbb{S}^*, \mathbb{Q}^*\}$, $\{\mathbb{S}, \mathbb{Q}\}$ by Equation \ref{eq-RSSPP}, \ref{eq_projector}\;
		Initialize $\mathbb{K}$, $\mathbb{W}$  by Equation \ref{eq_anchor_k}, \ref{eq-proto}\;
		\For{ $i \gets 0$ \KwTo $iter$ }{
			\For{ $t \gets 0$ \KwTo $\tau$ }{
				Compute $\bm{H}_{t}$, $\bm{H}_{tk}^{W}$ by Equation \ref{eq-log_map}\;
				Compute $p_{k,t}^{\bm{X}}$ by Equation \ref{eq_prob}\;
				Compute $\mathcal{L}^{ce}$, $\mathcal{L}^{mi}$ by Equation \ref{eq_ltim}\;			
			}
			Compute $\mathcal{L}$ by Equation \ref{eq_loss}\;
			Update 	$\mathbb{K}$, $\mathbb{W}$ by Equation \ref{eq_update} and \ref{eq-exp_map}\;
		}
		
		\KwOut{Query predictions are computed by Equation \ref{eq_score}\; }
		
		\caption{Pseudo-code for Classification}
		\label{algo}
	\end{algorithm}

	\textbf{Initialize Weights. } \label{sec_init} Empirically, we observe that it gains from initializing weights manually. Similar to the anchors, we calculate the prototypes of support features with weighted mean, and embed the class prototypes to OM:
	\begin{align}\label{eq-proto}
		\bm{W}_k  
		:=\operatorname{P_x} ( \frac{1}{k_S} \sum_{\bm{X}^{*}_S \in \mathbb{S}^{*}} \bm{X}^{*}_S \delta (y ({\bm{X}^{*}_S})=k) )
	\end{align}
	where $y ({\bm{X}^{*}_S)}$ is the label of $\bm{X}^{*}_S$, and $ \delta (y ({\bm{X}^{*}_S})=k) =1$ if and only if $y ({\bm{X}^{*}_S)}$  equals $k$.

	\textbf{Classification. }   The geodesic distance is difficult to compute and modern Euclidean transformation can not be applied directly. Since the tangent vector offers Euclidean properties, we transform the geodesic distance as the tangent vector $\bm{H}$, which is the shortest path from anchor $\bm{K}_t$ pointing towards $\bm{X}$ on OM by $\operatorname{Log}$. Then, we use $\operatorname{softmax}$ to calculate the distribution.
	
	Specially, following Equation \ref{eq-log_map}, we acquire the support and query tangent vectors $\bm{H}_{t}$ by $\bm{H}_{t}:=\operatorname{Log}_{\bm{K}_t} (\bm{X})$ where $\bm{X} \in \mathbb{S} \text{ or } \mathbb{Q} $ and weights tangent vectors $\bm{H}_{t}^{W_k}$ by $\bm{H}_{t}^{W_k}:=\operatorname{Log}_{\bm{K}_t} (\bm{W}_k)$ at each anchor $\bm{K}_t$ respectively. The distribution  over classes $\mathbb{P} (y=k|\bm{X},\bm{K}_t, \mathbb{W})$ in the tangent spaces is denoted as $p_{k,t}^{\bm{X}}$ :
	\begin{align}\label{eq_prob}
		p_{k,t}^{\bm{X}}:&=\frac{\exp (-\gamma\|\bm{H}_t - \bm{H}_{t}^{W_k}\|^2)}{\sum_{k'}\exp (-\gamma\|\bm{H}_t - \bm{H}^{W_{k'}}_{t}\|^2)} \notag \\
		& = \frac{\exp (-\gamma\|\operatorname{Log}_{\bm{K}_t} (\bm{X}) - \operatorname{Log}_{\bm{K}_t} (\bm{W}_k)\|^2)}{\sum_{k'}\exp (-\gamma\|\operatorname{Log}_{\bm{K}_t} (\bm{X}) - \operatorname{Log}_{\bm{K}_t} (\bm{W}_{k'})\|^2)} 
	\end{align}
	where $\gamma$ is a temperature parameter, and $\bm{X} \in \mathbb{S} \text{ or } \mathbb{Q} $.

	\begin{table*}[!htpb]
		\small
		\centering
		\begin{tabular}{lcccccc}
			&         &          & \multicolumn{2}{c}{ \textbf{\textit{mini}-ImageNet}} & \multicolumn{2}{c}{\textbf{\textit{tiered}-ImageNet}} \\

			Method & Type & Backbone & 1-shot          & 5-shot          & 1-shot           & 5-shot           \\
			
			\toprule

			TPN~\cite{liuLEARNINGPROPAGATELABELS2019} &Trans.& ResNet-12 &59.46& 75.64 &-& - \\
			TEAM~\cite{qiaoTransductiveEpisodicWiseAdaptive2019}& Trans. &ResNet-18& 60.07& 75.90 &- &- \\
			Transductive tuning~\cite{dhillonBaselineFewShotImage2020}& Trans.& ResNet-12 & 62.35  $\pm$  0.66 &74.53  $\pm$  0.54 &-& -\\
			MetaOpt~\cite{leeMetaLearningDifferentiableConvex2019} & Induc.&ResNet-12 &62.65 $\pm$ 0.61 &78.63 $\pm$ 0.46& 65.99 $\pm$ 0.72 &81.56 $\pm$ 0.53  \\
			DSN~\cite{simonAdaptiveSubspacesFewShot2020}& Induc.&ResNet-12 &64.60 $\pm$ 0.73&79.51 $\pm$ 0.50& 67.39 $\pm$ 0.82& 82.85 $\pm$ 0.56 \\	
			SimpleShot~\cite{wangSimpleShotRevisitingNearestNeighbor2019} &Induc.& ResNet-18& 63.10 $\pm$ 0.20& 79.92 $\pm$ 0.14& 69.68 $\pm$ 0.22 &84.56 $\pm$ 0.16  \\
			
			S2M2$_R$~\cite{manglaChartingRightManifold2020}&Induc.& ResNet-18 &64.06 $\pm$ 0.18& 80.58 $\pm$ 0.12 &  - & - \\

			Distill~\cite{tianRethinkingFewShotImage2020} & Induc.&ResNet-12 &64.82 $\pm$ 0.60& 82.14 $\pm$ 0.43& 71.52 $\pm$ 0.69& 86.03 $\pm$ 0.49 \\
			DeepEMD~\cite{zhangDeepEMDFewShotImage2020}& Induc. & ResNet-12& {65.91  $\pm$ 0.82} & 82.41 $\pm$ 0.56 & 71.16 $\pm$ 0.87 & 86.03 $\pm$ 0.58 \\

			LaplacianShot~\cite{zikoLaplacianRegularizedFewShot2020}&Trans.&ResNet-18&72.11  $\pm$  0.19 & 82.31  $\pm$  0.14 & 78.98 $\pm$  0.21 &86.39  $\pm$  0.16 \\
			
			TIM~\cite{boudiafTransductiveInformationMaximization2020}& Trans.& ResNet-18& 73.9 & 85.0& 79.9& 88.5 \\
			
			\midrule
			{Ours } &{Induc.}& {ResNet-18} & 63.98 $\pm$ 0.29 &{82.47 $\pm$ 0.44}&{70.50 $\pm$ 0.31}&{86.71 $\pm$ 0.49} \\
			\textbf{Ours} &\textbf{Trans.}& \textbf{ResNet-18} & \textbf{77.20 $\pm$ 0.36} &\textbf{87.11 $\pm$ 0.42}&\textbf{83.73 $\pm$ 0.36}&\textbf{90.46 $\pm$ 0.46}  \\
			
			\midrule

			LEO~\cite{rusuMETALEARNINGLATENTEMBEDDING2019}&Induc. & WRN28-10& 61.76  $\pm$  0.08 & 77.59  $\pm$  0.12& 66.33  $\pm$  0.05 &81.44  $\pm$  0.09\\
			CC+rot~\cite{gidaris2019boosting}& Induc. & WRN28-10& 62.93  $\pm$  0.45& 79.87  $\pm$  0.33& 70.53  $\pm$  0.51 &84.98  $\pm$  0.36\\
			MatchingNet~\cite{vinyals2016matching}&Induc. & WRN28-10 &64.03  $\pm$  0.20& 76.32  $\pm$  0.16 &-& -\\
			FEAT~\cite{yeFewShotLearningEmbedding2020}&Induc. & WRN28-10& 65.10  $\pm$  0.20 &81.11  $\pm$  0.14 &70.41  $\pm$  0.23& 84.38  $\pm$  0.16\\
			SimpleShot~\cite{wangSimpleShotRevisitingNearestNeighbor2019}&Induc.& WRN28-10 &65.87 $\pm$  0.20& 82.09  $\pm$  0.14& 70.90  $\pm$  0.22 &85.76  $\pm$  0.15\\
			
			S2M2$_R$~\cite{manglaChartingRightManifold2020}&Induc.& WRN28-10 &64.93 $\pm$ 0.18& 83.18 $\pm$ 0.11& 73.71 $\pm$ 0.22 &88.59 $\pm$ 0.14\\

			Transductive tuning~\cite{dhillonBaselineFewShotImage2020}& Trans.& WRN28-10 & 65.73  $\pm$  0.68 &78.40  $\pm$  0.52 &73.34  $\pm$  0.71 &85.50  $\pm$  0.50\\
			SIB~\cite{hu2020empirical}&Trans. & WRN28-10& 70.0  $\pm$  0.6 &79.2  $\pm$  0.4 &-& -\\
			BD-CSPN~\cite{liuPrototypeRectificationFewShot2020a}&Trans. &  WRN28-10& 70.31  $\pm$  0.93 &81.89  $\pm$  0.60& 78.74  $\pm$  0.95& 86.92  $\pm$  0.63\\
			LaplacianShot~\cite{boudiafTransductiveInformationMaximization2020} &Trans. & WRN28-10 & 74.86  $\pm$  0.19& 84.13  $\pm$  0.14& 80.18  $\pm$  0.21 &87.56 $\pm$  0.15\\
			IFSL~\cite{yueInterventionalFewShotLearning2020} &Trans. & WRN28-10 & 73.51  $\pm$  0.56& 83.21  $\pm$  0.33& 83.07  $\pm$   0.52 & 88.69 $\pm$  0.33\\
			TIM~\cite{boudiafTransductiveInformationMaximization2020}& Trans.& WRN28-10 & 77.8 & 87.4& 82.1& 89.8 \\
			
			\midrule
			
			{Ours } &{Induc.}& {WRN28-10} &{66.78 $\pm$ 0.30}&{85.29 $\pm$ 0.41}& {71.54 $\pm$ 0.29}&{87.79 $\pm$ 0.46} \\
			\textbf{Ours} &\textbf{Trans.}& \textbf{WRN28-10} &\textbf{80.64 $\pm$ 0.34 }&\textbf{89.39 $\pm$ 0.39}&\textbf{85.22$ \pm $ 0.34}&\textbf{91.35 $\pm$ 0.42} \\
			
			\bottomrule

		\end{tabular}

		\caption{ \textbf{Compared with the state-of-the-art on \textit{mini}-ImageNet and \textit{tiered}-ImageNet. } The results are 5-way few-shot classification scores averaged on 900 episodes for 5-shot and 4000 episodes for 1-shot, with 95\% confidence interval.  ``-'' signifies the result is available. ``Trans.'' and ``Indus.'' stand for the transductive and instructive settings. } 
		\label{table_mini_tiered}
		
	\end{table*}

	\subsection{Optimization on the Oblique Manifold}
	\textbf{Classification Loss }.  We follow TIM ~\cite{boudiafTransductiveInformationMaximization2020} and define the cross-entropy loss $\mathcal{L}^{ce} (\mathbb{W}, \bm{K}_t)$ with regard to the supervised support set, and the weighted mutual information $\mathcal{L}^{mi} (\mathbb{W}, \bm{K}_t)$  for
	the unlabeled query samples in the tangent spaces as:
	\small
	\begin{align}\label{eq_ltim}
		\mathcal{L}^{ce} (\mathbb{W}, \bm{K}_t) &: = -\lambda \frac{1}{|S|}\sum_{\bm{X} \in \mathbb{S}} \sum_{k=1}^{c}\delta (y ({\bm{X}})=k)\log (p_{k,t}^{\bm{X}}) \notag \\
		\mathcal{L}^{mi} (\mathbb{W}, \bm{K}_t) & :=  -\alpha \frac{1}{|Q|}\sum_{\bm{X} \in \mathbb{Q} }\sum_{k=1}^{c}p_{k,t}^{\bm{X}}\log (p_{k,t}^{\bm{X}}) \notag \\
		& + \sum_{k=1}^{c}\frac{1}{|Q|}\sum_{\bm{X} \in \mathbb{Q} }p_{k,t}^{\bm{X}} \log (\frac{1}{|Q|}\sum_{\bm{X} \in \mathbb{Q} }p_{k,t}^{\bm{X}})          
	\end{align}
	\normalsize
	with non-negative hyper-parameter $\alpha=\lambda=0.1$. 
	
	As aforementioned, the representation ability of $\operatorname{Log}$ decreases to points too far from anchors. It is not advisable to apply the same weight factors of $\mathcal{L}^{ce} (\mathbb{W},\bm{K}_t)$ and $\mathcal{L}^{mi} (\mathbb{W}, \bm{K}_t)$ over all the anchors. Intuitively, if $t$ is smaller, $\bm{K}_{t}$ will be closer to the support points. Then the support tangent vectors based $\bm{K}_{t}$ will be more credible, the large weight factor is suitable and vice versa. However, when $t=\frac{\tau}{2}$, $\bm{K}_{t}$ will be the global mean of all samples from both the support set and the query set. It conveys that the support tangent vectors at $\bm{K}_{t}$ are also reliable due to robust to the outliers, and then it requires the large weight factor. Based on the above observations, we propose the weight factor $\mu (t)$  over $t$ on $\mathcal{L}^{ce} (\mathbb{W}, \bm{K}_t)$ empirically:
	\small
	\begin{equation}\label{eq_ut} 
		\mu (t) := 
		\begin{cases}
			1 & \tau=0 \\		
			\frac{-t (2t-\tau)^2+\tau^3}{\tau^3} & \tau>0    
		\end{cases}
	\end{equation}
	\normalsize
	Finally, the predicts of query $s^{\bm{X}}$ are:   
	\small
	\begin{equation}\label{eq_score} 
		s^{\bm{X}} := \arg\max
		\begin{cases}
			p_{k,t}^{\bm{X}} & \tau=0 \\			
			\sum_t^\tau(1-\mu(t))p_{k,t}^{\bm{X}} & \tau>0    
		\end{cases} 
	\end{equation}
	\normalsize
	where $\bm{X} \in \mathbb{Q}$ and the classification loss $\mathcal{L}:=\mathcal{L} (X, W)$ is:
	\small
	\begin{align}\label{eq_loss} 
		\mathcal{L}& (\mathbb{W}, \mathbb{K}) \notag \\
		&:= \frac{\sum_{t=0}^\tau (\mu (t)\mathcal{L}^{ce} (\mathbb{W}, \bm{K}_t)+ (1-\mu (t))\mathcal{L}^{mi} (\mathbb{W}, \bm{K}_t))}{\sum_{t=0}^\tau \mu (t)} 
	\end{align}
	\normalsize

	\textbf{Optimization. }
	Both $\mathbb{W}$ and $\mathbb{K}$ have a similar optimization process. As an example, $\mathbb{W}$ is updated by Riemannian stochastic gradient descent  (RSGD)~\cite{becigneul2018riemannian,bonnabel2013stochastic} which updates parameters along the shortest path in manifold and avoiding leaving the manifold.  
	
	Specially, we firstly compute Euclidean gradient ${\nabla }_\mathbb{W} \mathcal{L} $ with respect to $\mathbb{W}$  by ${\nabla }_\mathbb{W} \mathcal{L} :=\frac{d}{d\mathbb{W}}\mathcal{L} (\operatorname{Log}_{\bm{K}}\bm{X}, \operatorname{Log}_{\bm{K}}\bm{W})$. Then, we project the Euclidean gradient ${\nabla }_\mathbb{W} \mathcal{L} $ to Riemannian gradient $\operatorname{G}_\mathbb{W}\mathcal{L}$ , which means the tangent vectors ${\nabla }_\mathbb{W} \mathcal{L} $ at $\mathbb{W}$ pointing to the updated parameters $\mathbb{W}^{ (k+1)}$. Referring to Equation \ref{eq-exp_map},  $\mathbb{W}^{ (k+1)}$ is computed by:
	\begin{align}\label{eq_update}
		\mathbb{W}^{ (k+1)} := \operatorname{Exp}_{\mathbb{W}^{ (k)}} -\lambda_r{\nabla }_{\mathbb{W}^{ (k)}} \mathcal{L} 
	\end{align}
	where $\lambda_r$ is the learning rate of RSGD.  
	The pseudo-code for classification on OM is shown in Algorithm \ref{algo}.

	\section{Experiments}

	\begin{table}[t]
		\centering
		\small
		\begin{tabular}{lccc|c}
			& & \multicolumn{2}{c}{ \textbf{CUB}} & {\textbf{M} $\to$  \textbf{C} } \\
			
			Method & Type  & 1-shot          & 5-shot            & 5-shot           \\		
			\toprule
			
			MAML~\cite{finn2017model} & Induc. & 68.42 & 83.47 & 51.34 \\
			MatchingNet~\cite{vinyals2016matching} & Induc & 73.49 & 84.45 &  53.07 \\
			ProtoNet~\cite{snellPrototypicalNetworksFewshot2017} &Induc & 72.99 & 86.64 & 62.02  \\
			RelationNet~\cite{sung2018learning} &Induc & 68.58 & 84.05 & 57.71 \\
			IFSL~\cite{yueInterventionalFewShotLearning2020}$^{\ddagger}$ &Induc & - & - & 60.05 \\
			Chen~\cite{chenCloserLookFewshot2019}& Induc & 67.02 & 83.58 &  65.57 \\
			SimpleShot ~\cite{wangSimpleShotRevisitingNearestNeighbor2019} & Induc & 70.28 & 86.37 & 65.63\\
			S2M2$_R$~\cite{manglaChartingRightManifold2020}& Induc. & 71.81 & 86.22 & 70.44 \\
			DeepEMD~\cite{zhangDeepEMDFewShotImage2020}$^{\dagger}$ &Induc & 75.65 & 88.69 & - \\

			IFSL~\cite{yueInterventionalFewShotLearning2020}$^{\ddagger}$ &Trans. & - & - & 62.07 \\
			LaplacianShot~\cite{zikoLaplacianRegularizedFewShot2020} & Trans. & 80.96 & 88.68 & 66.33 \\
			TIM~\cite{boudiafTransductiveInformationMaximization2020} & Trans. & 82.2 & 90.8 &71.0 \\
			
			\midrule

			{Ours}& Induc. &{78.24}&{92.15}&{72.47} \\

			\textbf{Ours}& Trans. &\textbf{85.87 }&\textbf{94.97}&\textbf{74.11} \\
			
			\bottomrule
			
		\end{tabular}

		\caption{Results for 5-shot and 1-shot CUB, and 5-shot for cross-domain on \textbf{\textit{mini}-ImageNet} $\to$  \textbf{CUB} with 5-way and backbone ResNet18.  ${\dagger}$ is based on  ResNet12 and  ${\ddagger}$ utilizes  ResNet10.}
		\label{table_cub}
		
	\end{table}

	\subsection{Implementation Details}
	\textbf{Backbone.} In all of our experiments, we take ResNet-18 and WRN28-10 as backbone.
	
	\textbf{Training.} Follow ~\cite{zikoLaplacianRegularizedFewShot2020,boudiafTransductiveInformationMaximization2020}, the backbone is trained with standard cross-entropy on the base classes in Euclidean space with L2-normalization activations. In this way, the features will fall on the oblique manifold naturally. The training epochs are 90, and the learning rate of SGD optimizer is 0.1 initially, which is divided by 10 at epochs 45 and epochs 66. The images are resized to 84 $\times$ 84. The batch size is 256 for ResNet-18 and 128 for WRN28-10. Same as ~\cite{zikoLaplacianRegularizedFewShot2020}, we apply data augmentation, \eg color jitter, random cropping, and random horizontal flipping and utilize label-smoothing with the parameter 0.1. 
	
	\textbf{Fine-tuning and Hyper-parameters. } The backbones are fixed during fine-tuning phase. We evaluate 4000 tasks for 5-way 1-shot and 900 tasks for 5-way 5-shot sampled from test datasets randomly and report the average accuracy along with 95\% confidence interval. To stay in line with TIM~\cite{boudiafTransductiveInformationMaximization2020},  we set $\alpha=0.1$, $\lambda=0.1$ in Equation \ref{eq_ltim} and $\gamma=7.5$ in Equation \ref{eq_prob}. Empirically, we set $q=11$ in Equation \ref{eq_mp}, $\tau=14$ for transductive inference, and $\tau=0$ in Equation \ref{eq_anchor_k} for inductive inference.  We use the RSGD~\cite{becigneul2018riemannian} optimizer with the learning rate $\lambda_r=0.1$ and the iterations for each task is 100.

	\begin{figure}[t]
		\centering
		
		\subfigure[Contour plot over $\tau$ and $p$]{
			\includegraphics[width=.225\textwidth]{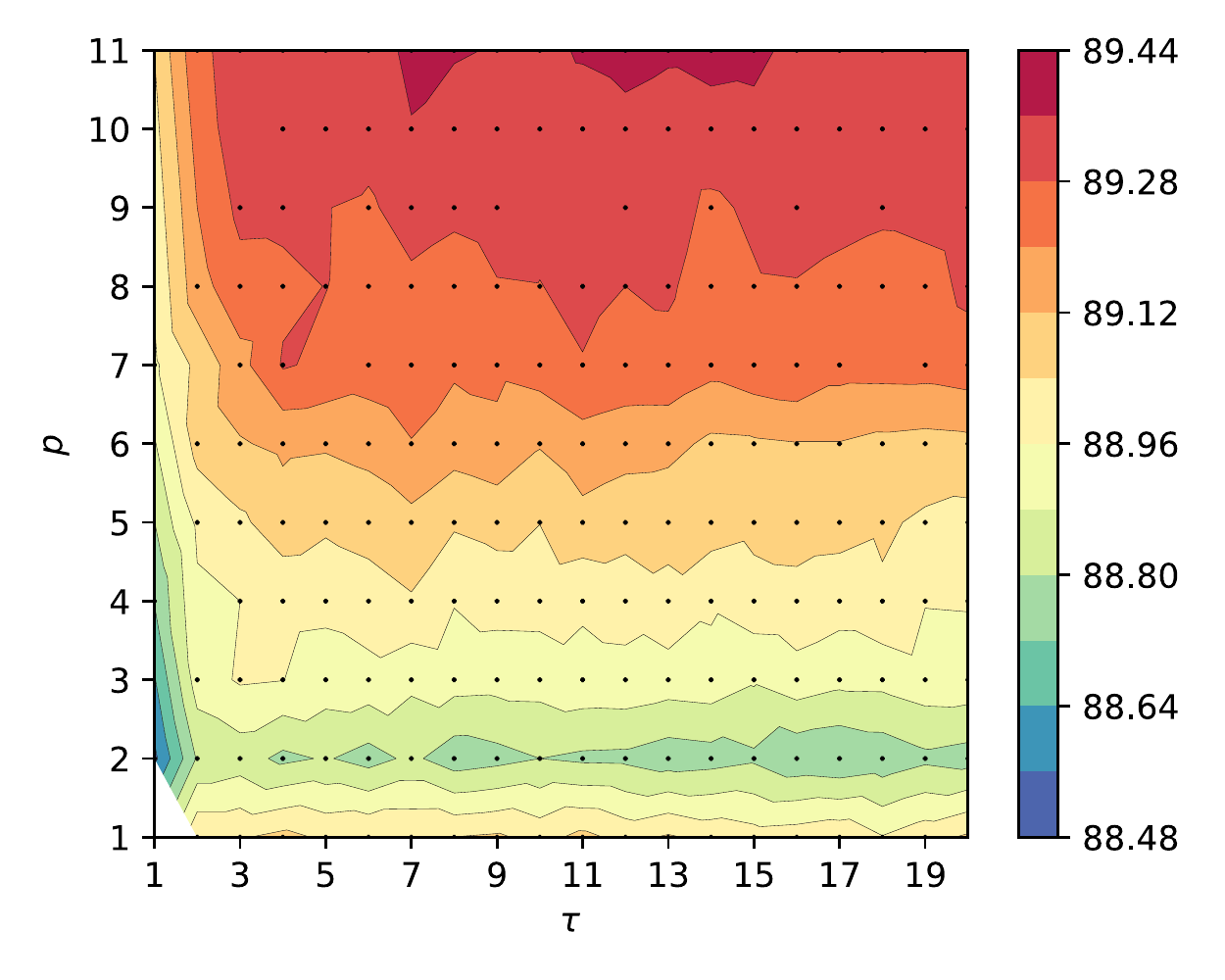}
			\label{fig_ablation_t_p}
		}
		\subfigure[Score lines graph over $\tau$ ]{
			\includegraphics[width=.22\textwidth]{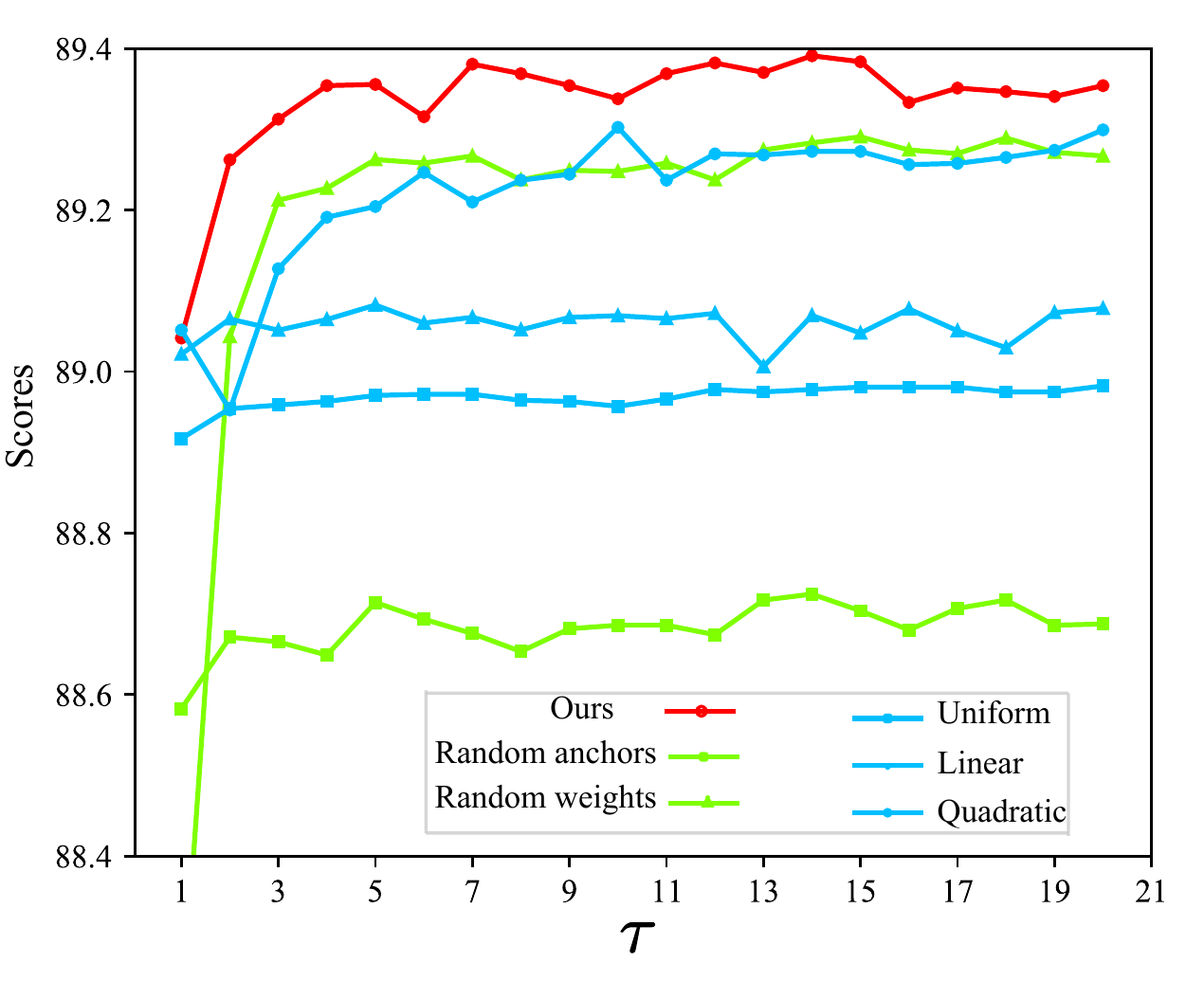}
			\label{fig_ablation_random}
		}

		\caption{Ablation study.  (a) Contour plot: visualization of classification scores over $\tau$ and $p$. Each point stands for a test with associated parameters.  (b) Score lines graph:the red line denotes our method, the chartreuse green lines denote studying on the initialized methods, and the sky blue lines mean researching the weight factor of loss. }
		
	\end{figure}

	\subsection{Datasets} 
	Three popular benchmark datasets are conducted: \textit{mini}-ImageNet~\cite{vinyals2016matching}, \textit{tiered}-ImageNet~\cite{ren2018meta}, and Caltech-UCSD Birds-200-2011  (CUB)~\cite{wah2011caltech}.   
	
	\textbf{\textit{mini}-ImageNet. } \textit{mini}-ImageNet is the subset of ImageNet datasets~\cite{russakovsky2015imagenet}. It contains 60,000 images totally with 84 $\times$ 84 size and  100 classes, which are divided into 64, 16, and 20 for training, validation and testing respectively. 
	
	\textbf{\textit{tiered}-ImageNet. } \textit{tiered}-ImageNet is also another one of ImageNet with 779,165 images and 84 $\times$ 84 size in total. It includes 34 classes and 608 sub-classes. The splits of training, validation, and testing are 20, 6, and 8 respectively, based on the high-level classes to enlarge the semantic gap between training and testing. 
	
	\textbf{Caltech-UCSD Birds-200-2011  (CUB). } CUB is used for fine-grained bird classification with a minor difference. It contains 200 classes and 11,788 images in all and the classes are divided into 100, 50, and 50 for training, validation, and testing respectively.

	\begin{table*}[!htpb]
		\centering
		 
		\begin{tabular}{lccccccccc}
			&         &     &     & \multicolumn{2}{c}{ \textbf{\textit{mini}-ImageNet}} & \multicolumn{2}{c}{\textbf{\textit{tiered}-ImageNet}}  & \multicolumn{2}{c}{\textbf{CUB}}  \\
			
			Riemannian &ODC  & RSSPP & Trans. & 1-shot          & 5-shot          & 1-shot           & 5-shot &1-shot           & 5-shot       \\
			\toprule
			
			Euclidean & \xmark & \xmark  & \xmark  & 65.25  & 82.17  & 70.19 & 85.51 & 76.80 & 90.31 \\  
			Euclidean &  \xmark & \cmark & \cmark & 66.39 & 83.33 &64.26 &84.19&81.25&91.86 \\
			L2-Euclidean &\xmark  & \xmark & \xmark & 67.61  &  82.24 &  66.50 & 84.00  & 77.30 & 90.10  \\
			L2-Euclidean & \xmark & \cmark & \cmark & 68.33 & 83.83 & 65.78 &  84.25 &81.22& 92.48 \\
			
			\midrule
			
			Oblique &  \cmark  & \xmark  & \xmark  & 66.00  & 84.34 & 71.29 & 87.27 & 77.98 & 91.51 \\  		 
			Oblique & \cmark  & \cmark &\xmark & 66.78 &85.29 & 71.54 & 87.79 & 78.24 & 92.15 \\
			Oblique & \cmark &  \xmark &\cmark & 80.22 & 88.71 & 84.70 & 91.20 & 85.01 & 94.37 \\

			\textbf{Ours} &\cmark  & \cmark & \cmark &\textbf{80.64} & \textbf{89.39} & \textbf{85.22} & \textbf{91.35} & \textbf{85.87} & \textbf{94.97} \\
			
			\bottomrule

		\end{tabular}
		
		\caption
		{The effectiveness of RSSPP, ODC and transductive on \textit{mini}-ImageNet, \textit{tiered}-ImageNet, and CUB, with backbone WRN28-10, WRN28-10 and Resnet18 respectively. }
		\label{table_ablastion}
		
	\end{table*}

	\subsection{Comparison with State-of-the-Art Methods}
	
	\textbf{Few-shot image classification. } From Table \ref{table_mini_tiered}, we can observe that our method outperforms the state-of-the-art in the transductive settings, and the gains are consistent with different network models and datasets. Note that our inductive model beats other inductive methods in 5-shot scenario. It shows that OM is more discriminative than Euclidean space and with more data the generalization ability of OM are enhanced rapidly.
	
	\textbf{Fine-grained image classification. } Table \ref{table_cub} reports the results of fine-grained 5-way 5-shot classification on CUB. Our transductive method outperforms the best state-of-the-art by about 3.6\% and 4.2\%  margins in 1-shot and 5-shot respectively. This shows that precise geodesic distance is more suitable than Euclidean distance to distinguish close features. 
	
	\textbf{Cross-domain (\textit{mini}-ImageNet $\to$ CUB). } Cross-domain few-shot classification is a challenging scenario. Following ~\cite{chenCloserLookFewshot2019}, we train Resnet-18 with  training set from \textit{mini}-ImageNet and validation set from CUB. The results in Table \ref{table_cub} (rightmost column) is evaluated on 5-way 5-shot from test set of CUB. We can observe that both our methods outperform others by about 3\% margins.

	\subsection{Ablation Study}

	\textbf{Effects of RSSPP, ODC, and transductive. } To check the effectiveness of  components of our method, we made some comparison experiences. Results are reported in Table \ref{table_ablastion}. In the first row, we replace the  oblique manifold with Euclidean space, remove RSSPP and transductive settings, \ie, $\tau=p=0$, and utilize the global average-polling directly. In this way, the model acts as transforming features by centering~\cite{wangSimpleShotRevisitingNearestNeighbor2019}. 
	The third and fourth rows show results of Euclidean with L2-normalization  (noted ``L2-Euclidean'') to compare the geodesic distance and Euclidean distance. We also carry out experiments to illustrate the efficacy of modules individually. The deference between the first four rows and the last four rows is: 1) The former are based on Euclidean distance while the latter are  based on geodesic distance. 2) Parameters of the former are updated in Euclidean space while parameters of the latter are updated along the surface of manifold. The results show that the scores gain about 0.5\% by RSSPP and 1\%-3\% by ODC in the transductive settings, and the ODC is better than Euclidean classifier.

	\textbf{Effects of ${\tau}$ and $p$. } We conduct an ablation study on the effect of   $p$ and ${\tau}$. Figure \ref{fig_ablation_t_p} is the contour plot that depicts how classification scores change with $p$ and ${\tau}$ in 5-shot settings on WRN28-10, where  $p$ ranges from 1 to 11, ${\tau}$ ranges from 1 to 20, and both are integers. It can be observed that: 1) The larger $p$, the better scores may be due to a better generalization and discriminative ability. 2) As ${\tau}$ increases, the higher scores. The reason is that more local approximate representations are utilized. However, the scores do not increase steadily. The explanation may be that the weight factor of single well-approximated representation decreases in the proportion of total representations according to Equation \ref{eq_loss}.

	\textbf{The initialization of anchors and weights. } We also study the effect of initialization of anchors and weights. The results are reported in Figure \ref{fig_ablation_random} with chartreuse green. We observe that anchors with random initialization perform the worst. It hardly catches the local approximate representation without considering the structure of manifold-features, and the parameters may be updated in the wrong direction. Random initialization of weights also leads to worse performance than our method by 0.15\% margin, which shows that it is necessary to initialize weights carefully.

	\textbf{The Weight factor of Loss.}  
	For multi-tangent spaces, suitable weight factors are required to determine the importance of scores according to the anchors, which engendered by the ratio of support points and query points. We propose another there functions of weight factors: 1) Uniform distribution, the factors of all terms are the same, \ie, $ \mu (t)=\frac{1}{2} $. 2) Linear function, the factors are reduced linearly,  \ie, $\mu (t) = 1 - \frac{t}{\tau}$. 3) Quadratic function when $t$ moves far, the factor is reduced sharply, \ie, $\mu (t) = 1 - ( \frac{t}{\tau})^2$. We conduct experiments to compare these factor functions, and the results are shown in Figure \ref{fig_ablation_random} with deep sky blue. Our method outperforms best over all $\tau$.

	\section{Conclusion}
	
	In this paper, we leverage the geodesic distance on the oblique manifold to solve the few-shot classification. The geodesic distance is transformed to vector in the tangent space and the modern machine learning tools can be utilized. We introduce a novel RSSPP for balancing generalization and discriminative ability, oblique distance-based classifier with new initial methods and a novel loss function for accurate classification. Finally, We validate the effectiveness on popular datasets. In the future, we will focus on how to integrate the local tangent space effectively or optimize the oblique manifold directly without resorting to the tangent space.
	
	\paragraph{Acknowledgement.} This work was supported in part by the Research Project of ZJU-League Research \& Development Center, Zhejiang Lab under
	Grant 2019KD0AB01.

	{\small
		\bibliographystyle{ieee_fullname}
		\bibliography{egbib}

\begin{thebibliography}{10}\itemsep=-1pt

\bibitem{absilJointDiagonalizationOblique2006}
P.~. {Absil} and K.~A. {Gallivan}.
\newblock Joint diagonalization on the oblique manifold for independent
  component analysis.
\newblock In {\em 2006 IEEE International Conference on Acoustics Speech and
  Signal Processing Proceedings}, volume~5, pages V--V, 2006.

\bibitem{becigneul2018riemannian}
Gary B{\'{e}}cigneul and Octavian{-}Eugen Ganea.
\newblock Riemannian adaptive optimization methods.
\newblock In {\em Proc. of ICLR}. OpenReview.net, 2019.

\bibitem{bonnabel2013stochastic}
Silvere Bonnabel.
\newblock Stochastic gradient descent on riemannian manifolds.
\newblock {\em IEEE Transactions on Automatic Control}, 58(9):2217--2229, 2013.

\bibitem{boudiafTransductiveInformationMaximization2020}
Malik Boudiaf, Imtiaz~Masud Ziko, J{\'{e}}r{\^{o}}me Rony, Jose Dolz, Pablo
  Piantanida, and Ismail~Ben Ayed.
\newblock Information maximization for few-shot learning.
\newblock In Hugo Larochelle, Marc'Aurelio Ranzato, Raia Hadsell,
  Maria{-}Florina Balcan, and Hsuan{-}Tien Lin, editors, {\em Advances in
  Neural Information Processing Systems 33: Annual Conference on Neural
  Information Processing Systems 2020, NeurIPS 2020, December 6-12, 2020,
  virtual}, 2020.

\bibitem{chakrabortyManifoldNetDeepNeural2020}
R. Chakraborty, J. Bouza, J. Manton, and B.~C. Vemuri.
\newblock Manifoldnet: A deep neural network for manifold-valued data
  withapplications.
\newblock {\em IEEE Transactions on Pattern Analysis and Machine Intelligence},
  pages 1--1, 2020.

\bibitem{chen2017sca}
Long Chen, Hanwang Zhang, Jun Xiao, Liqiang Nie, Jian Shao, Wei Liu, and
  Tat{-}Seng Chua.
\newblock {SCA-CNN:} spatial and channel-wise attention in convolutional
  networks for image captioning.
\newblock In {\em 2017 {IEEE} Conference on Computer Vision and Pattern
  Recognition, {CVPR} 2017, Honolulu, HI, USA, July 21-26, 2017}, pages
  6298--6306. {IEEE} Computer Society, 2017.

\bibitem{chenCloserLookFewshot2019}
Wei{-}Yu Chen, Yen{-}Cheng Liu, Zsolt Kira, Yu{-}Chiang~Frank Wang, and
  Jia{-}Bin Huang.
\newblock A closer look at few-shot classification.
\newblock In {\em Proc. of ICLR}. OpenReview.net, 2019.

\bibitem{dhillonBaselineFewShotImage2020}
Guneet~Singh Dhillon, Pratik Chaudhari, Avinash Ravichandran, and Stefano
  Soatto.
\newblock A baseline for few-shot image classification.
\newblock In {\em Proc. of ICLR}. OpenReview.net, 2020.

\bibitem{dvornikDiversityCooperationEnsemble2019}
Nikita Dvornik, Julien Mairal, and Cordelia Schmid.
\newblock Diversity with cooperation: Ensemble methods for few-shot
  classification.
\newblock In {\em 2019 {IEEE/CVF} International Conference on Computer Vision,
  {ICCV} 2019, Seoul, Korea (South), October 27 - November 2, 2019}, pages
  3722--3730. {IEEE}, 2019.

\bibitem{dvornikSelectingRelevantFeatures2020}
Nikita Dvornik, Cordelia Schmid, and Julien Mairal.
\newblock Selecting relevant features from a multi-domain representation for
  few-shot classification.
\newblock In {\em European Conference on Computer Vision}, pages 769--786.
  Springer, 2020.

\bibitem{eisenhart1997riemannian}
Luther~Pfahler Eisenhart.
\newblock {\em Riemannian geometry}, volume~51.
\newblock Princeton university press, 1997.

\bibitem{finn2017model}
Chelsea Finn, Pieter Abbeel, and Sergey Levine.
\newblock Model-agnostic meta-learning for fast adaptation of deep networks.
\newblock In Doina Precup and Yee~Whye Teh, editors, {\em Proceedings of the
  34th International Conference on Machine Learning, {ICML} 2017, Sydney, NSW,
  Australia, 6-11 August 2017}, volume~70 of {\em Proceedings of Machine
  Learning Research}, pages 1126--1135. {PMLR}, 2017.

\bibitem{fu2019dual}
Jun Fu, Jing Liu, Haijie Tian, Yong Li, Yongjun Bao, Zhiwei Fang, and Hanqing
  Lu.
\newblock Dual attention network for scene segmentation.
\newblock In {\em {IEEE} Conference on Computer Vision and Pattern Recognition,
  {CVPR} 2019, Long Beach, CA, USA, June 16-20, 2019}, pages 3146--3154.
  Computer Vision Foundation / {IEEE}, 2019.

\bibitem{gidaris2019boosting}
Spyros Gidaris, Andrei Bursuc, Nikos Komodakis, Patrick P{\'{e}}rez, and
  Matthieu Cord.
\newblock Boosting few-shot visual learning with self-supervision.
\newblock In {\em 2019 {IEEE/CVF} International Conference on Computer Vision,
  {ICCV} 2019, Seoul, Korea (South), October 27 - November 2, 2019}, pages
  8058--8067. {IEEE}, 2019.

\bibitem{gongGeodesicFlowKernel2012}
Boqing Gong, Yuan Shi, Fei Sha, and Kristen Grauman.
\newblock Geodesic flow kernel for unsupervised domain adaptation.
\newblock In {\em 2012 {IEEE} Conference on Computer Vision and Pattern
  Recognition, Providence, RI, USA, June 16-21, 2012}, pages 2066--2073. {IEEE}
  Computer Society, 2012.

\bibitem{gopalanDomainAdaptationObject2011}
Raghuraman Gopalan, Ruonan Li, and Rama Chellappa.
\newblock Domain adaptation for object recognition: An unsupervised approach.
\newblock In Dimitris~N. Metaxas, Long Quan, Alberto Sanfeliu, and Luc~Van
  Gool, editors, {\em {IEEE} International Conference on Computer Vision,
  {ICCV} 2011, Barcelona, Spain, November 6-13, 2011}, pages 999--1006. {IEEE}
  Computer Society, 2011.

\bibitem{grove1973conjugatec}
Karsten Grove and Hermann Karcher.
\newblock How to conjugatec 1-close group actions.
\newblock {\em Mathematische Zeitschrift}, 132(1):11--20, 1973.

\bibitem{harandiGraphEmbeddingDiscriminant2011}
Mehrtash~Tafazzoli Harandi, Conrad Sanderson, Sareh~Abolahrari Shirazi, and
  Brian~C. Lovell.
\newblock Graph embedding discriminant analysis on grassmannian manifolds for
  improved image set matching.
\newblock In {\em The 24th {IEEE} Conference on Computer Vision and Pattern
  Recognition, {CVPR} 2011, Colorado Springs, CO, USA, 20-25 June 2011}, pages
  2705--2712. {IEEE} Computer Society, 2011.

\bibitem{he2015spatial}
Kaiming He, Xiangyu Zhang, Shaoqing Ren, and Jian Sun.
\newblock Spatial pyramid pooling in deep convolutional networks for visual
  recognition.
\newblock {\em IEEE transactions on pattern analysis and machine intelligence},
  37(9):1904--1916, 2015.

\bibitem{he2016deep}
Kaiming He, Xiangyu Zhang, Shaoqing Ren, and Jian Sun.
\newblock Deep residual learning for image recognition.
\newblock In {\em 2016 {IEEE} Conference on Computer Vision and Pattern
  Recognition, {CVPR} 2016, Las Vegas, NV, USA, June 27-30, 2016}, pages
  770--778. {IEEE} Computer Society, 2016.

\bibitem{hinton2006reducing}
Geoffrey~E Hinton and Ruslan~R Salakhutdinov.
\newblock Reducing the dimensionality of data with neural networks.
\newblock {\em Science}, 313(5786):504--507, 2006.

\bibitem{hotelling1933analysis}
Harold Hotelling.
\newblock Analysis of a complex of statistical variables into principal
  components.
\newblock {\em Journal of educational psychology}, 24(6):417, 1933.

\bibitem{hu2019local}
Han Hu, Zheng Zhang, Zhenda Xie, and Stephen Lin.
\newblock Local relation networks for image recognition.
\newblock In {\em 2019 {IEEE/CVF} International Conference on Computer Vision,
  {ICCV} 2019, Seoul, Korea (South), October 27 - November 2, 2019}, pages
  3463--3472. {IEEE}, 2019.

\bibitem{hu2020empirical}
Shell~Xu Hu, Pablo~Garcia Moreno, Yang Xiao, Xi Shen, Guillaume Obozinski,
  Neil~D. Lawrence, and Andreas~C. Damianou.
\newblock Empirical bayes transductive meta-learning with synthetic gradients.
\newblock In {\em Proc. of ICLR}. OpenReview.net, 2020.

\bibitem{huang2017riemannian}
Zhiwu Huang and Luc~Van Gool.
\newblock A riemannian network for {SPD} matrix learning.
\newblock In Satinder~P. Singh and Shaul Markovitch, editors, {\em Proc. of
  AAAI}, pages 2036--2042. {AAAI} Press, 2017.

\bibitem{huangProjectionMetricLearning2015}
Zhiwu Huang, Ruiping Wang, Shiguang Shan, and Xilin Chen.
\newblock Projection metric learning on grassmann manifold with application to
  video based face recognition.
\newblock In {\em {IEEE} Conference on Computer Vision and Pattern Recognition,
  {CVPR} 2015, Boston, MA, USA, June 7-12, 2015}, pages 140--149. {IEEE}
  Computer Society, 2015.

\bibitem{kimFewshotVisualReasoning2020}
Youngsung Kim, Jinwoo Shin, Eunho Yang, and Sung~Ju Hwang.
\newblock Few-shot visual reasoning with meta-analogical contrastive learning.
\newblock In Hugo Larochelle, Marc'Aurelio Ranzato, Raia Hadsell,
  Maria{-}Florina Balcan, and Hsuan{-}Tien Lin, editors, {\em Advances in
  Neural Information Processing Systems 33: Annual Conference on Neural
  Information Processing Systems 2020, NeurIPS 2020, December 6-12, 2020,
  virtual}, 2020.

\bibitem{leeMetaLearningDifferentiableConvex2019}
Kwonjoon Lee, Subhransu Maji, Avinash Ravichandran, and Stefano Soatto.
\newblock Meta-learning with differentiable convex optimization.
\newblock In {\em {IEEE} Conference on Computer Vision and Pattern Recognition,
  {CVPR} 2019, Long Beach, CA, USA, June 16-20, 2019}, pages 10657--10665.
  Computer Vision Foundation / {IEEE}, 2019.

\bibitem{li2019learning}
Xinzhe Li, Qianru Sun, Yaoyao Liu, Qin Zhou, Shibao Zheng, Tat{-}Seng Chua, and
  Bernt Schiele.
\newblock Learning to self-train for semi-supervised few-shot classification.
\newblock In Hanna~M. Wallach, Hugo Larochelle, Alina Beygelzimer, Florence
  d'Alch{\'{e}}{-}Buc, Emily~B. Fox, and Roman Garnett, editors, {\em Advances
  in Neural Information Processing Systems 32: Annual Conference on Neural
  Information Processing Systems 2019, NeurIPS 2019, December 8-14, 2019,
  Vancouver, BC, Canada}, pages 10276--10286, 2019.

\bibitem{lifchitz2019dense}
Yann Lifchitz, Yannis Avrithis, Sylvaine Picard, and Andrei Bursuc.
\newblock Dense classification and implanting for few-shot learning.
\newblock In {\em {IEEE} Conference on Computer Vision and Pattern Recognition,
  {CVPR} 2019, Long Beach, CA, USA, June 16-20, 2019}, pages 9258--9267.
  Computer Vision Foundation / {IEEE}, 2019.

\bibitem{liuPrototypeRectificationFewShot2020a}
Jinlu Liu, Liang Song, and Yongqiang Qin.
\newblock Prototype rectification for few-shot learning.
\newblock In {\em {{ECCV}}}, 2020.

\bibitem{liuLEARNINGPROPAGATELABELS2019}
Yanbin Liu, Juho Lee, Minseop Park, Saehoon Kim, Eunho Yang, Sung~Ju Hwang, and
  Yi Yang.
\newblock Learning to propagate labels: Transductive propagation network for
  few-shot learning.
\newblock In {\em Proc. of ICLR}. OpenReview.net, 2019.

\bibitem{liuEnsembleEpochwiseEmpirical2020}
Yaoyao Liu, Bernt Schiele, and Qianru Sun.
\newblock An ensemble of epoch-wise empirical bayes for few-shot learning.
\newblock In {\em European Conference on Computer Vision}, pages 404--421.
  Springer, 2020.

\bibitem{manglaChartingRightManifold2020}
Puneet Mangla, Mayank Singh, Abhishek Sinha, Nupur Kumari, Vineeth~N
  Balasubramanian, and Balaji Krishnamurthy.
\newblock Charting the right manifold: Manifold mixup for few-shot learning.
\newblock In {\em {{WACV}}}, pages 2207--2216, {Snowmass Village, CO, USA},
  2020. {IEEE}.

\bibitem{montiGeometricDeepLearning2017}
Federico Monti, Davide Boscaini, Jonathan Masci, Emanuele Rodol{\`{a}}, Jan
  Svoboda, and Michael~M. Bronstein.
\newblock Geometric deep learning on graphs and manifolds using mixture model
  cnns.
\newblock In {\em 2017 {IEEE} Conference on Computer Vision and Pattern
  Recognition, {CVPR} 2017, Honolulu, HI, USA, July 21-26, 2017}, pages
  5425--5434. {IEEE} Computer Society, 2017.

\bibitem{nguyenNeuralNetworkBased2019}
Xuan~Son Nguyen, Luc Brun, Olivier Lezoray, and S{\'{e}}bastien Bougleux.
\newblock A neural network based on {SPD} manifold learning for skeleton-based
  hand gesture recognition.
\newblock In {\em {IEEE} Conference on Computer Vision and Pattern Recognition,
  {CVPR} 2019, Long Beach, CA, USA, June 16-20, 2019}, pages 12036--12045.
  Computer Vision Foundation / {IEEE}, 2019.

\bibitem{park2019meta}
Eunbyung Park and Junier~B. Oliva.
\newblock Meta-curvature.
\newblock In Hanna~M. Wallach, Hugo Larochelle, Alina Beygelzimer, Florence
  d'Alch{\'{e}}{-}Buc, Emily~B. Fox, and Roman Garnett, editors, {\em Advances
  in Neural Information Processing Systems 32: Annual Conference on Neural
  Information Processing Systems 2019, NeurIPS 2019, December 8-14, 2019,
  Vancouver, BC, Canada}, pages 3309--3319, 2019.

\bibitem{patacchiolaBayesianMetaLearningFewShot2020}
Massimiliano Patacchiola, Jack Turner, Elliot~J. Crowley, Michael
  O\textquotesingle~Boyle, and Amos~J Storkey.
\newblock Bayesian meta-learning for the few-shot setting via deep kernels.
\newblock In H. Larochelle, M. Ranzato, R. Hadsell, M.~F. Balcan, and H. Lin,
  editors, {\em Advances in Neural Information Processing Systems}, volume~33,
  pages 16108--16118. Curran Associates, Inc., 2020.

\bibitem{pautratOnlineInvarianceSelection2020}
R{\'e}mi Pautrat, Viktor Larsson, Martin~R. Oswald, and Marc Pollefeys.
\newblock Online invariance selection for local feature descriptors.
\newblock In Andrea Vedaldi, Horst Bischof, Thomas Brox, and Jan-Michael Frahm,
  editors, {\em ECCV 2020}, volume 12347, pages 707--724, {Cham}, 2020.
  {Springer International Publishing}.

\bibitem{qiaoTransductiveEpisodicWiseAdaptive2019}
Limeng Qiao, Yemin Shi, Jia Li, Yonghong Tian, Tiejun Huang, and Yaowei Wang.
\newblock Transductive episodic-wise adaptive metric for few-shot learning.
\newblock In {\em 2019 {IEEE/CVF} International Conference on Computer Vision,
  {ICCV} 2019, Seoul, Korea (South), October 27 - November 2, 2019}, pages
  3602--3611. {IEEE}, 2019.

\bibitem{rajeswaran2019meta}
Aravind Rajeswaran, Chelsea Finn, Sham~M. Kakade, and Sergey Levine.
\newblock Meta-learning with implicit gradients.
\newblock In Hanna~M. Wallach, Hugo Larochelle, Alina Beygelzimer, Florence
  d'Alch{\'{e}}{-}Buc, Emily~B. Fox, and Roman Garnett, editors, {\em Advances
  in Neural Information Processing Systems 32: Annual Conference on Neural
  Information Processing Systems 2019, NeurIPS 2019, December 8-14, 2019,
  Vancouver, BC, Canada}, pages 113--124, 2019.

\bibitem{ren2018meta}
Mengye Ren, Eleni Triantafillou, Sachin Ravi, Jake Snell, Kevin Swersky,
  Joshua~B. Tenenbaum, Hugo Larochelle, and Richard~S. Zemel.
\newblock Meta-learning for semi-supervised few-shot classification.
\newblock In {\em Proc. of ICLR}. OpenReview.net, 2018.

\bibitem{russakovsky2015imagenet}
Olga Russakovsky, Jia Deng, Hao Su, Jonathan Krause, Sanjeev Satheesh, Sean Ma,
  Zhiheng Huang, Andrej Karpathy, Aditya Khosla, Michael Bernstein, et~al.
\newblock Imagenet large scale visual recognition challenge.
\newblock {\em International journal of computer vision}, 115(3):211--252,
  2015.

\bibitem{rusuMETALEARNINGLATENTEMBEDDING2019}
Andrei~A. Rusu, Dushyant Rao, Jakub Sygnowski, Oriol Vinyals, Razvan Pascanu,
  Simon Osindero, and Raia Hadsell.
\newblock Meta-learning with latent embedding optimization.
\newblock In {\em Proc. of ICLR}. OpenReview.net, 2019.

\bibitem{simonAdaptiveSubspacesFewShot2020}
Christian Simon, Piotr Koniusz, Richard Nock, and Mehrtash Harandi.
\newblock Adaptive subspaces for few-shot learning.
\newblock In {\em 2020 {IEEE/CVF} Conference on Computer Vision and Pattern
  Recognition, {CVPR} 2020, Seattle, WA, USA, June 13-19, 2020}, pages
  4135--4144. {IEEE}, 2020.

\bibitem{snellPrototypicalNetworksFewshot2017}
Jake Snell, Kevin Swersky, and Richard~S. Zemel.
\newblock Prototypical networks for few-shot learning.
\newblock In Isabelle Guyon, Ulrike von Luxburg, Samy Bengio, Hanna~M. Wallach,
  Rob Fergus, S.~V.~N. Vishwanathan, and Roman Garnett, editors, {\em Advances
  in Neural Information Processing Systems 30: Annual Conference on Neural
  Information Processing Systems 2017, December 4-9, 2017, Long Beach, CA,
  {USA}}, pages 4077--4087, 2017.

\bibitem{souzaInterfaceGrassmannManifolds2020}
L.~S. Souza, N. Sogi, B.~B. Gatto, T. Kobayashi, and K. Fukui.
\newblock An interface between grassmann manifolds and vector spaces.
\newblock In {\em {{CVPR Workshop}}}, pages 3695--3704, 2020.

\bibitem{sung2018learning}
Flood Sung, Yongxin Yang, Li Zhang, Tao Xiang, Philip H.~S. Torr, and
  Timothy~M. Hospedales.
\newblock Learning to compare: Relation network for few-shot learning.
\newblock In {\em 2018 {IEEE} Conference on Computer Vision and Pattern
  Recognition, {CVPR} 2018, Salt Lake City, UT, USA, June 18-22, 2018}, pages
  1199--1208. {IEEE} Computer Society, 2018.

\bibitem{tan2019mnasnet}
Mingxing Tan, Bo Chen, Ruoming Pang, Vijay Vasudevan, Mark Sandler, Andrew
  Howard, and Quoc~V. Le.
\newblock Mnasnet: Platform-aware neural architecture search for mobile.
\newblock In {\em {IEEE} Conference on Computer Vision and Pattern Recognition,
  {CVPR} 2019, Long Beach, CA, USA, June 16-20, 2019}, pages 2820--2828.
  Computer Vision Foundation / {IEEE}, 2019.

\bibitem{tenenbaum2000global}
Joshua~B Tenenbaum, Vin De~Silva, and John~C Langford.
\newblock A global geometric framework for nonlinear dimensionality reduction.
\newblock {\em Science}, 290(5500):2319--2323, 2000.

\bibitem{tianRethinkingFewShotImage2020}
Yonglong Tian, Yue Wang, Dilip Krishnan, Joshua~B. Tenenbaum, and Phillip
  Isola.
\newblock Rethinking few-shot image classification: A good embedding is all you
  need?
\newblock In {\em {{ECCV}}}, 2020.

\bibitem{trendafilovMultimodeProcrustesProblem2002}
Nickolay~T. Trendafilov and Ross~A. Lippert.
\newblock The multimode procrustes problem.
\newblock {\em Linear Algebra and its Applications}, 349(1):245--264, 2002.

\bibitem{vinyals2016matching}
Oriol Vinyals, Charles Blundell, Tim Lillicrap, Koray Kavukcuoglu, and Daan
  Wierstra.
\newblock Matching networks for one shot learning.
\newblock In Daniel~D. Lee, Masashi Sugiyama, Ulrike von Luxburg, Isabelle
  Guyon, and Roman Garnett, editors, {\em Advances in Neural Information
  Processing Systems 29: Annual Conference on Neural Information Processing
  Systems 2016, December 5-10, 2016, Barcelona, Spain}, pages 3630--3638, 2016.

\bibitem{wah2011caltech}
C. Wah, S. Branson, P. Welinder, P. Perona, and S. Belongie.
\newblock {The Caltech-UCSD Birds-200-2011 Dataset}.
\newblock Technical Report CNS-TR-2011-001, California Institute of Technology,
  2011.

\bibitem{wangGraphEmbeddingMultiKernel2021}
R. Wang, X.-J. Wu, and J. Kittler.
\newblock Graph embedding multi-kernel metric learning for image set
  classification with grassmannian manifold-valued features.
\newblock {\em IEEE Transactions on Multimedia}, 23:228--242, 2021.

\bibitem{Wang_2018_CVPR}
Xiaolong Wang, Ross~B. Girshick, Abhinav Gupta, and Kaiming He.
\newblock Non-local neural networks.
\newblock In {\em 2018 {IEEE} Conference on Computer Vision and Pattern
  Recognition, {CVPR} 2018, Salt Lake City, UT, USA, June 18-22, 2018}, pages
  7794--7803. {IEEE} Computer Society, 2018.

\bibitem{wangSimpleShotRevisitingNearestNeighbor2019}
Yan Wang, Wei-Lun Chao, Kilian~Q. Weinberger, and Laurens {van der Maaten}.
\newblock Simpleshot: Revisiting nearest-neighbor classification for few-shot
  learning.
\newblock In {\em {{arXiv:1911.04623}}}, 2019.

\bibitem{weiLearningDiscriminativeGeodesic2018}
Jianre Wei, Jian Liang, Ran He, and Jinfeng Yang.
\newblock Learning discriminative geodesic flow kernel for unsupervised domain
  adaptation.
\newblock In {\em {{ICME}}}, pages 1--6, {San Diego, CA}, 2018. {IEEE}.

\bibitem{woo2018cbam}
Sanghyun Woo, Jongchan Park, Joon-Young Lee, and In~So Kweon.
\newblock Cbam: Convolutional block attention module.
\newblock In {\em ECCV}, pages 3--19, 2018.

\bibitem{yangDPGNDistributionPropagation2020}
Ling Yang, Liangliang Li, Zilun Zhang, Xinyu Zhou, Erjin Zhou, and Yu Liu.
\newblock {DPGN:} distribution propagation graph network for few-shot learning.
\newblock In {\em 2020 {IEEE/CVF} Conference on Computer Vision and Pattern
  Recognition, {CVPR} 2020, Seattle, WA, USA, June 13-19, 2020}, pages
  13387--13396. {IEEE}, 2020.

\bibitem{yeFewShotLearningEmbedding2020}
Han{-}Jia Ye, Hexiang Hu, De{-}Chuan Zhan, and Fei Sha.
\newblock Few-shot learning via embedding adaptation with set-to-set functions.
\newblock In {\em 2020 {IEEE/CVF} Conference on Computer Vision and Pattern
  Recognition, {CVPR} 2020, Seattle, WA, USA, June 13-19, 2020}, pages
  8805--8814. {IEEE}, 2020.

\bibitem{yueInterventionalFewShotLearning2020}
Zhongqi Yue, Hanwang Zhang, Qianru Sun, and Xian-Sheng Hua.
\newblock Interventional few-shot learning.
\newblock In {\em Advances in Neural Information Processing Systems},
  volume~33, 2020.

\bibitem{zhangDeepEMDFewShotImage2020}
Chi Zhang, Yujun Cai, Guosheng Lin, and Chunhua Shen.
\newblock Deepemd: Few-shot image classification with differentiable earth
  mover's distance and structured classifiers.
\newblock In {\em 2020 {IEEE/CVF} Conference on Computer Vision and Pattern
  Recognition, {CVPR} 2020, Seattle, WA, USA, June 13-19, 2020}, pages
  12200--12210. {IEEE}, 2020.

\bibitem{zhangFewShotLearningSaliencyGuided2019}
Hongguang Zhang, Jing Zhang, and Piotr Koniusz.
\newblock Few-shot learning via saliency-guided hallucination of samples.
\newblock In {\em {IEEE} Conference on Computer Vision and Pattern Recognition,
  {CVPR} 2019, Long Beach, CA, USA, June 16-20, 2019}, pages 2770--2779.
  Computer Vision Foundation / {IEEE}, 2019.

\bibitem{zhaoExploringSelfAttentionImage2020}
Hengshuang Zhao, Jiaya Jia, and Vladlen Koltun.
\newblock Exploring self-attention for image recognition.
\newblock In {\em 2020 {IEEE/CVF} Conference on Computer Vision and Pattern
  Recognition, {CVPR} 2020, Seattle, WA, USA, June 13-19, 2020}, pages
  10073--10082. {IEEE}, 2020.

\bibitem{zikoLaplacianRegularizedFewShot2020}
Imtiaz Ziko, Jose Dolz, Eric Granger, and Ismail~Ben Ayed.
\newblock Laplacian regularized few-shot learning.
\newblock In {\em International Conference on Machine Learning}, pages
  11660--11670. PMLR, 2020.

\end{thebibliography}
	}
	
\end{document}